\documentclass{article}
\usepackage{graphicx} % Required for inserting images
\usepackage{amsmath}
\usepackage{cite}
\usepackage[accepted]{synsml2023}
\usepackage{hyperref}
\usepackage{listings}

\begin{document}

\twocolumn[
\synsmltitle{Open Source Infrastructure for Differentiable Density Functional Theory}]

\begin{synsmlauthorlist}
\synsmlauthor{Advika Vidhyadhiraja}{comp}
\synsmlauthor{Arun Pa Thiagarajan}{comp}
\synsmlauthor{Shang Zhu}{yyy}
\synsmlauthor{Venkat Viswanathan}{yyy}
\synsmlauthor{Bharath Ramsundar}{comp}
\synsmlcorrespondingauthor{Bharath Ramsundar}{bharath@deepforestsci.com}
%\synsmlauthor{}{sch}
%\synsmlauthor{}{sch}
\synsmlaffiliation{yyy}{Department of Mechanical Engineering, Carnegie Mellon University, Pennsylvania, USA}
\synsmlaffiliation{comp}{Deep Forest Sciences, California, USA}
\end{synsmlauthorlist}
\printAffiliationsAndNotice

\begin{abstract}
Learning exchange correlation functionals, used in quantum chemistry calculations, from data has become increasingly important in recent years, but training such a functional requires sophisticated software infrastructure. For this reason, we build open source infrastructure to train neural exchange correlation functionals. We aim to standardize the processing pipeline by adapting state-of-the-art techniques from work done by multiple groups. We have open sourced the model in the DeepChem\cite{deepchembook} library to provide a platform for additional research on differentiable quantum chemistry methods. %We use a fully differentiable three dimensional quantum chemistry library to build a model and layers to train a functional that can be used for a heterogeneous data set. 
\end{abstract}

\section{Introduction}

Density-functional theory (DFT) is used to calculate the electronic structure of atoms, molecules, and solids. Its objective is to use the fundamental laws of quantum mechanics to quantitatively comprehend the properties of materials. There are serious scaling limitations to traditional methods used to approximate solutions to the Schr\"{o}dinger equation of $N$ interacting electrons moving in an external potential. Whereas in DFT, instead of the high-dimensional many-body wave function, the density $n(r)$ is a function of three spatial coordinates. The many-body electronic ground state can be described using single-particle equations and an effective potential thanks to the Kohn-Sham theory \cite{PhysRev.140.A1133}. 

The effective potential is made up of three parts: the exchange-correlation (XC) potential, which accounts for many-body effects, the Hartree potential, which describes the electrostatic electron-electron interaction, and the ionic potential resulting from the atomic cores \cite{KURTH2005395, PEDERSON2015153}. Mathematically, the energy contributors in DFT can be represented as

\begin{align}
E_{total} &= E_{\rm kin} + E_{\rm el} + E_{\rm xc}  
\end{align}

The kinetic energy term is calculated using a fictitious non-interacting system. The second term captures the electrostatic interactions between electrons and nuclear cores. A potential energy surface is derived using the Born-Oppenheimer approximation to account for the electrostatic repulsion between the nuclear cores \cite{Voss_2022}. The most commonly used and simplest class of XC functionals are the local-density approximations (LDA) which mandate that $E_{\textrm{xc}}$ depends only on $n(r)$ and not on its derivatives. Functionals such as the LDA class are traditional approximate forms derived by humans and are widely used due to their accuracy\cite{doi:10.1146/annurev.pc.46.100195.003413} .

\subsection{DeepChem and Differentiable Physics}

DeepChem is an open source python library for scientific machine learning and deep learning on molecular and quantum datasets \cite{Bharath_Peastman_Amacbride_nvtrang91_2021}. Deepchem provides a framework to solve difficult scientific problems in areas such as drug discovery, energy calculations and biotech \cite{C7SC02664A}. It does so by specifying that scientific calculations must be broken down into workflows built out of underlying primitives such as data loaders, featurizers, data splitters, learned models, metrics and hyperparameter tuners. This systematic design allows DeepChem to be applicable to a wide variety of applications. For example, DeepChem has enabled large scale benchmarking for molecular machine learning through the MoleculeNet benchmark suite \cite{C7SC02664A}, protein-ligand interaction modeling \cite{gomes2017atomic}, generative modeling of molecules \cite{frey2022fastflows}, and more.

DeepChem aims to support open source differentiable programming infrastructure for scientific machine learning, but this effort is a work in progress \cite{ramsundar2021differentiable}. This research program is broadly known as differentiable physics \cite{ramsundar2021differentiable}. An important application of differentiable physics is to use neural architectures to accelerate the solution of differential equations. In this work, we aim to solve the self-consistent Kohn Sham equations to calculate electron density $n(r)$. Typically, solution methods for self-consistent calculations can be slow. Differentiable techniques enable rapid calculations by introducing rich neural approximation schemes. %Lagrangian and Hamiltonian neural networks can be used to model complex systems and solve differential equations.

\subsection{Differentiable DFT Methods}

The incorporation of machine learning methods into DFT has been going on for over a decade \cite{2022NatRP...4..357P}.
Our model has been derived from XCNN \cite{Kasim_2021}. In this method, the xc-functional is learned using a deep neural network and hybridized with a traditional xc-functional. One of the biggest strengths of this approach, is that it is generalizable to a wide range of systems since the neural architecture does not depend on a special physical system \cite{Kasim_2021}. Similarly, another advance in this field was made by the DM21 functional \cite{doi:10.1126/science.abj6511}, which was computed using fictional systems having fractional charge and spin constraints in order to avoid errors encountered by traditional xc-functionals. These errors are observed for charge densities involving mobile charges and spins\cite{doi:10.1126/science.abj6511}.

\subsection{Standardizing Differentiable DFT Workflows}

We aim to standardize ("deepchemize" \cite{Bharath_Peastman_Amacbride_nvtrang91_2021}) the computation performed by the differentiable DFT model, by implementing the model using the workflow structure implemented by other DeepChem models. To verify correctness, we run experiments and compare results with the original XCNN model \cite{Kasim_2021}. By standardizing the process of training a neural network exchange correlation functional, we aim to make it simpler to experiment with new differentiable DFT architectures and enable larger systematic computations. 

%While hybridizing the neural network XC functional with a traditional functional with tunable weights. 
\lstinline{XCModel} (the XCNN implementation in DeepChem), can be used with different loss functions, metrics and machine learned models present in DeepChem, providing a flexible framework for xc-functional design. This flexibility of the XCModel can be used to incorporate more global density information to approximate a more accurate exchange correlation functional.  For example, we are currently planning on introducing fractional constraints similar to the DM21 functional to the XCModel. This project would yield a functional that is a combination of DM21 and XCNN. The functional would be trained using a neural network and a fully differentiable quantum chemistry library while also being trained on data points that contain fictional systems with fractional charges and spins.

\section{Implementation}
\subsection{Dataset}
\begin{figure}
\includegraphics[scale=0.5]{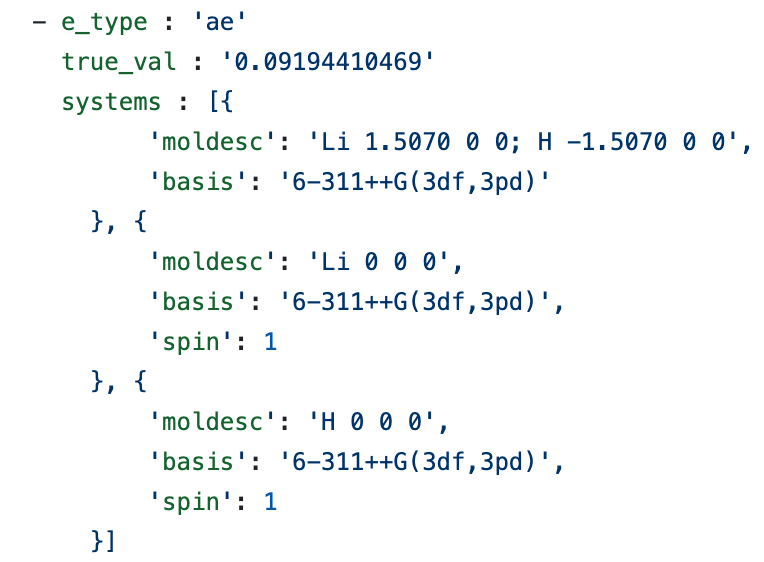}
\caption{Training data is provided in a Yaml format as displayed above. Input data is processed with the \lstinline{DFTYamlLoader} class.} 
\label{fig:Example Data}
\end{figure}

There are four types of data points used to train \lstinline{XCModel}: atomization energy (AE) calculations, ionization potential (IP) calculations, density profile regulations and density matrix calculations. The ground truth values for the first two types are obtained from NIST databases \cite{Kasim_2021}, and the rest are calculated by performing CCSD calculations (using PYSCF \cite{sun2018pyscf}). Users do not need to enter the equations used to calculate the total energy for AE and IP data points. In \cite{Kasim_2021}, the training dataset consists of all four types of data points. However, we have only used AE, IP and Density matrix calculations thus far. We have implemented a \lstinline{DFTYamlLoader}class in DeepChem that loads and prepares data, and featurizes the data into standard molecular objects. We use a specific format to build the yaml file. Each molecule is known as an ``entry" object , and contains multiple ``system" objects. Each system contains information on the molecule description, basis set, charge, and spin number. The true values are either numbers or .npy files. For predictions, the true values do not need to be entered. An example of a data point can be seen in Figure~\ref{fig:Example Data}

\subsection{Layers}
\begin{figure}
\includegraphics[scale=0.35]{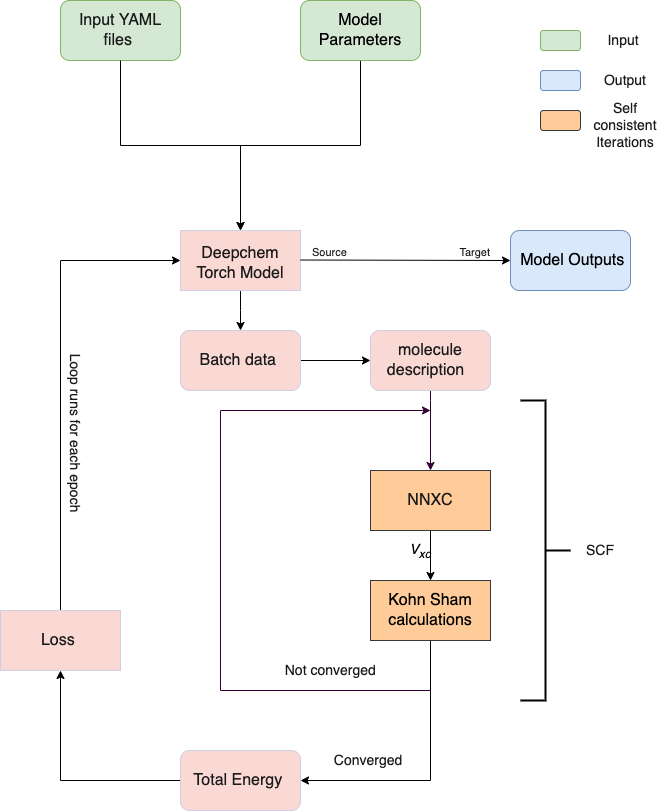}
\caption{Schematics of the exchange correlation \lstinline{XCModel}. The yaml files are loaded using the \lstinline{DFTYamlLoader} and featurized into mol objects using the DFT data classes. The model parameters consist of the PyTorch model used to train the functional and the choice of loss function. The forward method initializes the Neural Network LDA layer (NNXC), and hybridizes the functional with a traditional LDA functional. The hybrid xc is used to solve the Kohn-Sham equations. Once the self consistent iterations converge, the total energy of the data point is calculated and used to calculate the loss. The trained model can be used with DeepChem functions such as evaluate and predict with different metrics \cite{Kasim_2021, Bharath_Peastman_Amacbride_nvtrang91_2021}}. 
\label{fig:ModelFlowchart}
\end{figure}

A layer can be defined as a function that transforms a tensor into another. The first layer in the \lstinline{XCModel} is the Neural Network XC (NNXC) layer, where the exchange correlation functional is trained based on a pre-defined traditional class of functionals. Currently we have implemented an NNLDA and NNPBE layer. The electron and spin densities are transformed to be the input of the neural network. We plan to implement Meta-GGA based layers in the future. The output from this layer is used in the \lstinline{HybridXC} layer, which computes XC energy by summing XC energy computed from libxc (with any conventional DFT functional) and the trainable neural network with tunable weights. This layers corresponds to equation 2 from \cite{Kasim_2021}
\begin{align}
E_{nnLDA}[n] &=  \alpha E_{\rm LDA(xc)}[n] + \beta \int  n(r)f(n,\varepsilon) dr 
\end{align}
The output from the \lstinline{HybridXC} layer is used to calculate the XC potential which is used to solve the self consistent iterations. This process is carried out in the SCF (Self Consistent Field) layer. The SCF iterations are done using DQC's fully differentiable KS modules \cite{DQC}. This layer can also be used to perform dft calculations without any NNXC, i.e, any traditional xc functional. The outputs from this layer are the electron densities. The gradient propagation of the self-consistency cycle is done using an implicit gradient calculation instead of linear mixing using xitorch \cite{Kasim_2021}. These layers are used in the \lstinline{XCModel}. An ordinary feed forward neural network (implemented in PyTorch) can be used to train the XC functional. The torch model can also be used with various loss functions and metrics present in DeepChem . %All the layers and the torch model are written based on / derived from the XCNN library.\cite{XCNN, Kasim_2021}
In our experiments, we use a $L^2$ loss to train the model and mean absolute error (MAE) as the metric. A schematic of the implementation can be seen in Figure~\ref{fig:ModelFlowchart}. 

\section{Results and Observations}
Table~\ref{tab:result_table} is a comparison between MAE values for two different test datasets; when computed using various trained/ traditional functionals. The test datasets we use are: ionization potential for the atoms H-Ar, and atomization energy for 16 Hydrocarbons. The exact molecules can be found in the supplementary material of \cite{Kasim_2021}. Results indicate that the DeepChem implementation slightly trails the performance of the original implementation from \cite{Kasim_2021}, but we anticipate that this gap will close once we complete the implementation. The training and testing for these experiments can be run on a 16GB RAM CPU system. However, for testing on larger molecules and datasets, a GPU system would likely be required. 

\begin{table}[h!]
\centering
\begin{tabular}{ |p{3cm}|p{2cm}|p{2cm}| }
 \hline
 Calculations & IP 18 & AE 16 HC
 \\
 \hline
 LDA   & 24.6    &48.7 \\
 XCNN-LDA-IP &   15.2  & 25.4   \\
 DC-XCNN-LDA-IP & 24.2 &  28.8\\
 \hline
\end{tabular}
\caption{MAE scores in kcal/mol for two test datasets}
\label{tab:result_table}
\end{table}

\subsection{Hydrogen Dissociation  }

\begin{figure}    
\includegraphics[scale=0.35]{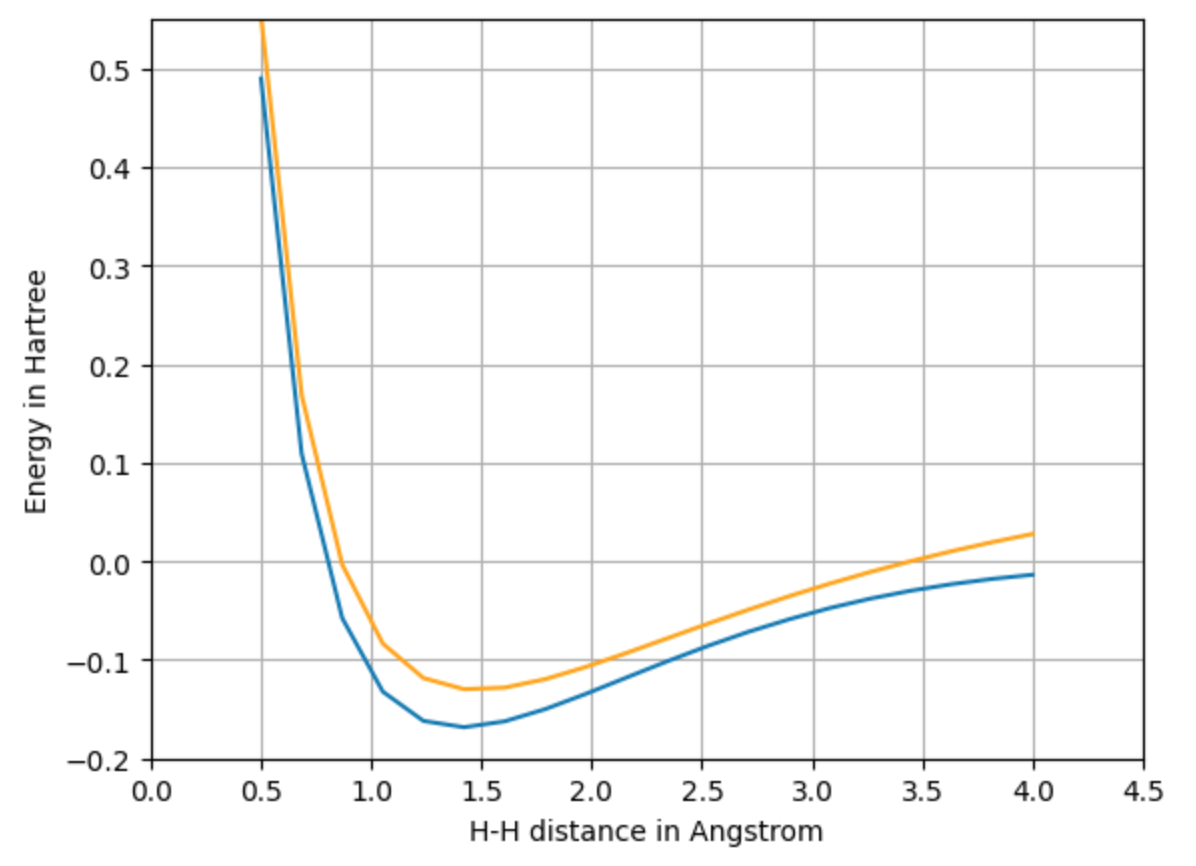}
\caption{Hydrogen Dissociation Curve. The orange graph represents the exact values\cite{XCNN} of the energy required for H2 to dissociate at different bond lengths. The blue graph represents the predicted energies using the trained model. We have used 20 data points for this calculation. }  
\label{fig:Dissociation Curve}
\end{figure}

In Fig~\ref{fig:Dissociation Curve}, we plot the dissociation energy for Hydrogen at different points and compare the values with CCSD values. The computed dissociation curve computed by \lstinline{XCModel} closely tracks the exact values.

\section{Conclusion}
From Table~\ref{tab:result_table}, we observe that the MAE deviations from our model are slightly more accurate compared to a traditional LDA functional, despite not being as accurate as XCNN \cite{Kasim_2021}. We can also conclude from the hydrogen dissociation diagram that predicted values are quite close in value to the exact values. Since \lstinline{XCModel} is flexible with its parameters, we have also experimented with using various loss functions, metrics, and PyTorch models to train the functional. In the near future, we plan on implementing a few features to further increase accuracy of the DFT calculations while also making the model more versatile. These features include fractional constraints on fictional systems, and layers based on PBE and Meta-PBE. We anticipate that the open DeepChem implementation will enable additional rapid experimentation with differentiable xc-functional architectures.

\section{Impact Statement}
%As discussed, introducing deep learning and neural networks to quantum chemistry methods, increases approximation accuracy which leads to more accurate prediction of ground state energies. Although this model is capable of making predictions using heterogeneous data, training the model on a specific type of data set may further increase prediction accuracy. Besides using a specific data set, the choice of traditional functional layer used to train the model can be quite useful as well. For example, organic and organic anionic molecules work best with a hybrid functional such as the Meta-PBE.\cite{C6CP06163J} Training the model on a Meta-GGA/PBE layer would lead to better predictions. Organic and organic anionic molecules are present in multiple parts of the human body such as liver, brain tissues, urate, bile salts and a lot of transportation factors. They are also used in anti-biotics.\cite{HAGOS20071} 
Differentiable DFT software infrastructure could enable systematic construction of more accurate exchange correlation functions, enabling DFT calculations to have greater impact in materials design and biotechnology applications where existing functionals lack accuracy today.

\bibliographystyle{synsml2023}
\bibliography{ref1}

\begin{thebibliography}{17}
\providecommand{\natexlab}[1]{#1}
\providecommand{\url}[1]{\texttt{#1}}
\expandafter\ifx\csname urlstyle\endcsname\relax
  \providecommand{\doi}[1]{doi: #1}\else
  \providecommand{\doi}{doi: \begingroup \urlstyle{rm}\Url}\fi

\bibitem[Frey et~al.(2022)Frey, Gadepally, and Ramsundar]{frey2022fastflows}
Frey, N.~C., Gadepally, V., and Ramsundar, B.
\newblock Fastflows: Flow-based models for molecular graph generation.
\newblock \emph{arXiv preprint arXiv:2201.12419}, 2022.

\bibitem[Gomes et~al.(2017)Gomes, Ramsundar, Feinberg, and
  Pande]{gomes2017atomic}
Gomes, J., Ramsundar, B., Feinberg, E.~N., and Pande, V.~S.
\newblock Atomic convolutional networks for predicting protein-ligand binding
  affinity.
\newblock \emph{arXiv preprint arXiv:1703.10603}, 2017.

\bibitem[Kasim(2021)]{XCNN}
Kasim, M.
\newblock Xcnn.
\newblock \url{https://github.com/mfkasim1/xcnn}, 2021.

\bibitem[Kasim \& Sofroniew(2021)Kasim and Sofroniew]{DQC}
Kasim, M. and Sofroniew, N.
\newblock Diffqc/dqc.
\newblock \url{https://github.com/diffqc/dqc}, 2021.

\bibitem[Kasim \& Vinko(2021)Kasim and Vinko]{Kasim_2021}
Kasim, M. and Vinko, S.
\newblock Learning the exchange-correlation functional from nature with fully
  differentiable density functional theory.
\newblock \emph{Physical Review Letters}, 127\penalty0 (12), sep 2021.
\newblock \doi{10.1103/physrevlett.127.126403}.
\newblock URL \url{https://doi.org/10.1103%2Fphysrevlett.127.126403}.

\bibitem[Kirkpatrick et~al.(2021)Kirkpatrick, McMorrow, Turban, Gaunt, Spencer,
  Matthews, Obika, Thiry, Fortunato, Pfau, Castellanos, Petersen, Nelson,
  Kohli, Mori-Sánchez, Hassabis, and Cohen]{doi:10.1126/science.abj6511}
Kirkpatrick, J., McMorrow, B., Turban, D. H.~P., Gaunt, A.~L., Spencer, J.~S.,
  Matthews, A. G. D.~G., Obika, A., Thiry, L., Fortunato, M., Pfau, D.,
  Castellanos, L.~R., Petersen, S., Nelson, A. W.~R., Kohli, P., Mori-Sánchez,
  P., Hassabis, D., and Cohen, A.~J.
\newblock Pushing the frontiers of density functionals by solving the
  fractional electron problem.
\newblock \emph{Science}, 374\penalty0 (6573):\penalty0 1385--1389, 2021.
\newblock \doi{10.1126/science.abj6511}.
\newblock URL \url{https://www.science.org/doi/abs/10.1126/science.abj6511}.

\bibitem[Kohn \& Sham(1965)Kohn and Sham]{PhysRev.140.A1133}
Kohn, W. and Sham, L.~J.
\newblock Self-consistent equations including exchange and correlation effects.
\newblock \emph{Phys. Rev.}, 140:\penalty0 A1133--A1138, Nov 1965.
\newblock \doi{10.1103/PhysRev.140.A1133}.
\newblock URL \url{https://link.aps.org/doi/10.1103/PhysRev.140.A1133}.

\bibitem[Kurth et~al.(2005)Kurth, Marques, and Gross]{KURTH2005395}
Kurth, S., Marques, M., and Gross, E.
\newblock Density-functional theory.
\newblock pp.\  395--402, 2005.
\newblock \doi{https://doi.org/10.1016/B0-12-369401-9/00445-9}.
\newblock URL
  \url{https://www.sciencedirect.com/science/article/pii/B0123694019004459}.

\bibitem[Parr \& Yang(1995)Parr and
  Yang]{doi:10.1146/annurev.pc.46.100195.003413}
Parr, R.~G. and Yang, W.
\newblock Density-functional theory of the electronic structure of molecules.
\newblock \emph{Annual Review of Physical Chemistry}, 46\penalty0 (1):\penalty0
  701--728, 1995.
\newblock \doi{10.1146/annurev.pc.46.100195.003413}.
\newblock URL \url{https://doi.org/10.1146/annurev.pc.46.100195.003413}.
\newblock PMID: 24341393.

\bibitem[Pederson \& Baruah(2015)Pederson and Baruah]{PEDERSON2015153}
Pederson, M.~R. and Baruah, T.
\newblock Chapter eight - self-interaction corrections within the
  fermi-orbital-based formalism.
\newblock 64:\penalty0 153--180, 2015.
\newblock ISSN 1049-250X.
\newblock \doi{https://doi.org/10.1016/bs.aamop.2015.06.005}.
\newblock URL
  \url{https://www.sciencedirect.com/science/article/pii/S1049250X15000087}.

\bibitem[{Pederson} et~al.(2022){Pederson}, {Kalita}, and
  {Burke}]{2022NatRP...4..357P}
{Pederson}, R., {Kalita}, B., and {Burke}, K.
\newblock {Machine learning and density functional theory}.
\newblock \emph{Nature Reviews Physics}, 4\penalty0 (6):\penalty0 357--358, May
  2022.
\newblock \doi{10.1038/s42254-022-00470-2}.

\bibitem[Ramsundar et~al.(2021{\natexlab{a}})Ramsundar, Peastman, Amacbride,
  and nvtrang91]{Bharath_Peastman_Amacbride_nvtrang91_2021}
Ramsundar, Peastman, Amacbride, and nvtrang91.
\newblock Making deepchem a better framework for ai-driven science, Apr
  2021{\natexlab{a}}.
\newblock URL
  \url{https://forum.deepchem.io/t/making-deepchem-a-better-framework-for-ai-driven-science/431}.

\bibitem[Ramsundar et~al.(2019)Ramsundar, Eastman, Walters, Pande, Leswing, and
  Wu]{deepchembook}
Ramsundar, B., Eastman, P., Walters, P., Pande, V., Leswing, K., and Wu, Z.
\newblock \emph{Deep Learning for the Life Sciences}.
\newblock O'Reilly Media, 2019.
\newblock
  \url{https://www.amazon.com/Deep-Learning-Life-Sciences-Microscopy/dp/1492039837}.

\bibitem[Ramsundar et~al.(2021{\natexlab{b}})Ramsundar, Krishnamurthy, and
  Viswanathan]{ramsundar2021differentiable}
Ramsundar, B., Krishnamurthy, D., and Viswanathan, V.
\newblock Differentiable physics: A position piece, 2021{\natexlab{b}}.

\bibitem[Sun et~al.(2018)Sun, Berkelbach, Blunt, Booth, Guo, Li, Liu, McClain,
  Sayfutyarova, Sharma, et~al.]{sun2018pyscf}
Sun, Q., Berkelbach, T.~C., Blunt, N.~S., Booth, G.~H., Guo, S., Li, Z., Liu,
  J., McClain, J.~D., Sayfutyarova, E.~R., Sharma, S., et~al.
\newblock Pyscf: the python-based simulations of chemistry framework.
\newblock \emph{Wiley Interdisciplinary Reviews: Computational Molecular
  Science}, 8\penalty0 (1):\penalty0 e1340, 2018.

\bibitem[Voss(2022)]{Voss_2022}
Voss, J.
\newblock Exchange-correlation functionals, 2022.
\newblock URL \url{https://stanford.edu/~vossj/slac/project/xc-functionals/}.

\bibitem[Wu et~al.(2018)Wu, Ramsundar, Feinberg, Gomes, Geniesse, Pappu,
  Leswing, and Pande]{C7SC02664A}
Wu, Z., Ramsundar, B., Feinberg, E., Gomes, J., Geniesse, C., Pappu, A.~S.,
  Leswing, K., and Pande, V.
\newblock Moleculenet: a benchmark for molecular machine learning.
\newblock \emph{Chem. Sci.}, 9:\penalty0 513--530, 2018.
\newblock \doi{10.1039/C7SC02664A}.
\newblock URL \url{http://dx.doi.org/10.1039/C7SC02664A}.

\end{thebibliography}

\end{document}